\documentclass[preprint,12pt]{elsarticle}
\usepackage{amssymb}
\usepackage{epsfig} 
\usepackage{graphicx}
\usepackage{subfigure}
\usepackage{xcolor}
\usepackage{gensymb}
\usepackage{amsmath}
\usepackage{hyperref}
\usepackage{bm}
\usepackage{float}
\usepackage{multirow}
\usepackage{caption}
\usepackage{graphicx}
\usepackage{algorithm}
\usepackage{mathtools}
\usepackage{algorithmic}
\usepackage{float}
\usepackage{url}
\usepackage{hyperref}
\usepackage{adjustbox}
\usepackage[a4paper, total={170mm,257mm}, left=15mm,top=15mm]{geometry}
\usepackage{amsmath,amssymb,amsfonts}
\usepackage{array}
\usepackage[mathscr]{euscript}
\usepackage{tabularx}
\begin{document}
\begin{frontmatter}
\title{Sanitizing Manufacturing Dataset Labels Using Vision-Language Models}

\author[inst1]{Nazanin Mahjourian\thanks{Corresponding Author}}
\ead{mahjouri@mtu.edu}

\author[inst1]{Vinh Nguyen}\ead{vinhn@mtu.edu}

\affiliation[inst1]{organization={Department of Mechanical Engineering - Engineering Mechanics},
            addressline={Michigan Technological University}, 
            city={Houghton},
            postcode={49931}, 
            state={MI},
            country={USA}}

\begin{abstract}

The success of machine learning models in industrial applications is heavily dependent on the quality of the datasets used to train the models. However, large-scale datasets, specially those constructed from crowd-sourcing and web-scraping, often suffer from label noise, inconsistencies, and errors, which can negatively impact model performance. This problem is particularly pronounced in manufacturing domains, where obtaining high-quality labels is costly and time-consuming. This paper introduces Vision-Language Sanitization and Refinement (VLSR), which is a vision-language-based framework for label sanitization and refinement in multi-label manufacturing image datasets. This method embeds both images and their associated textual labels into a shared semantic space leveraging the CLIP vision-language model. Then two key tasks are addressed in this process by computing the cosine similarity between embeddings. First, label sanitization is performed to identify irrelevant, misspelled, or semantically weak labels, and surface the most semantically aligned label for each image by comparing image-label pairs using cosine similarity between image and label embeddings. Second, the method applies density-based clustering on text embeddings, followed by iterative cluster merging, to group semantically similar labels, such as those differing only in casing, phrasing, synonymy, misspellings, or unnecessary specificity into unified label groups. The Factorynet dataset, which includes noisy labels from both human annotations and web-scraped sources, is employed to evaluate the effectiveness of the proposed framework. Experimental results demonstrate that the VLSR framework successfully identifies problematic labels and improves label consistency. This method enables a significant reduction in label vocabulary through clustering, which ultimately enhances the dataset's quality for training robust machine learning models in industrial applications with minimal human intervention. Therefore, this work presents a solution for dataset curation in multi-label manufacturing scenarios where label noise is prevalent.
  
\end{abstract}




\begin{keyword}
Vision-Language models \sep Multi-modal learning \sep Label Noise \sep Web-Scraping \sep Label Cleaning \sep Dataset Sanitization\sep Embedding Clustering
\end{keyword}

\end{frontmatter}


\section{Introduction}

In recent years, the expansion of artificial intelligence (AI) has revolutionized numerous domains from autonomous driving and robotics to industrial automation\cite{mohammadi2025gps,ziad2024knowledge, jandaghi2023motion,borazjani2025multi}. A growing body of research has demonstrated the effectiveness of AI in addressing a wide range of challenges in manufacturing, including predictive maintenance, defect detection, quality control, and process optimization \cite{hosseinzadeh2024minimizing, safari2022dependency,RezvaniBoroujeni2025, ziad2024pyramid}. One of the key factors for making AI to be successful in these applications is the quality of datasets used to train models. Many real-world applications rely heavily on high-quality datasets for tasks including object detection, segmentation, and scene understanding \cite{Drost_2017_ICCV}. A well-annotated and large-scale dataset is critical for robust model performance. Specifically, performance can suffer because inconsistency and errors in the labels in a large dataset can make it harder for algorithm to learn patterns \cite{lee2021surveydatacleaningmethods}. Real-world datasets can be susceptible to label noise, which refers to errors or inconsistencies in the labeling of data, such as incorrect, ambiguous, or conflicting labels \cite{frenay2014comprehensive}.

Label noise comes from multiple different sources, particularly when performing labeling manually. This is due to factors such as insufficient or poor-quality information, perceptual errors, and variability between humans conducting labeling. It can also come from communication problems, data encoding errors, or creating labels from non-experts \cite{6685834}. Crowd-sourcing is another method of labeling by assigning it to many non-experts, usually through an online platform. It is necessary to collect annotations from multiple annotators for a single instance, because non-experts are prone to making mistakes in crowd-sourcing tasks \cite{10.1145/1401890.1401965}. Once multiple labels are collected, they need to be combined into one final label using a process called consensus \cite{tang2011semi}. However, different experts might label the same instance in various methods, use different terms, or introduce spelling errors. Furthermore, differences in terminology, labeling styles, or even spelling errors from multiple annotators can introduce additional label noise into the dataset. Unfortunately, it is also expensive and challenging to obtain reliable labels using crowd-sourcing method \cite{khetan2017learning}.

To address the high costs of capturing and labeling large datasets by humans, modern approaches increasingly rely on web scraping \cite{Chen_2015_ICCV,5518767}. While this technique can accelerate data collection, it often introduces errors into the labeling process \cite{li2024multi}. For instance, the returned image of web scraping data using the label "bicycle" may return images that include a scene with multiple objects or cluttered backgrounds. A caption for a group photo might mention "bicycle" since a bicycle is in the background, but the image's focus isn't a bicycle. This will lead to confusion for the object detection model to learn the pattern of a bicycle image. Some images may even include symbolic representations or drawings of bicycle, which are not suitable for input data for a model in a real-world scenario. The scraped image can include irrelevant or broad context. It can include misspellings, synonyms, or entirely incorrect terms. As another example, a "bicycle" might be labeled "bike" inconsistently across the dataset. In addition, level of detail in labels may vary. One image might have a specific label like "mountain bike", while another is broadly labeled "bicycle", even though both are bicycles. It can also introduce class imbalance to the dataset, because certain categories might dominate the scraped data, skewing the distribution and making it harder for models to learn underrepresented classes. Such mismatches can hinder a classifier's ability to learn accurate patterns or lead to misleading associations \cite{chen2015webly,yao2019towards,divvala2014learning}.

Thus, there are many different approaches to address the problem of learning with noisy labels \cite{10206599}. Active label cleaning \cite{bernhardt2022active} is a data-driven method which selects samples for re-annotation by ranking instances based on estimated label correctness and difficulty. This not only is costly but also limits their cleaning ability. Sample selection method is another method to learn with noisy label \cite{NEURIPS2018_a19744e2,xia2021sample} by prioritizing cleanly labeled data, aiming to reduce the impact of noisy labeled samples on model performance. Noise-robust learning methods \cite{wang2019symmetric,liu2020early} also prevent the network overfitting to incorrect labels using noise-tolerant loss functions, early learning dynamics, and regularization to improve model performance. While these methods may increase the performance of the classifiers, they do not directly correct mislabeled data within the dataset.

While those traditional methods have been widely used, there has been limited research to explore the potential of foundation models in tackling noisy labels and enhancing data quality. There are only a few examples that demonstrate the use of these models in addressing  the challenges of noisy labels. Large Language Models (LLMs) have recently emerged as a powerful tool for addressing the data quality issues, because of their knowledge of semantics and natural language understanding. IterClean \cite{ni2024iterclean} is an iterative data cleaning framework that combines the data labeling with iterative cleaning steps such as error detection and repair using LLMs. The human-LLM collaborative framework \cite{wang2024human} uses human annotators to re-annotate the data using the guidance provided by LLMs. Crowd-LLM \cite{li2024comparative} integrates LLM labels with crowd-sourcing methods to enhance the quality of aggregated labels. However these methods are only relying on analyzing the text-only input, which is not useful for analyzing the quality of matching the label with its corresponding image pair. 

Vision-language models are another powerful tools that are being considered to increase the quality of the datasets by utilizing the alignment of textual and visual features. However, only a limited number of studies have explored their use for label cleaning. This lack of knowledge highlights a potential gap for using these powerful models to solve the real-world manufacturing problems. DeFT \cite{wei2024vision} uses positive and negative textual prompts for each class to detect noisy labels. CLIPCleaner \cite{feng2024clipcleaner} selects clean samples for Learning with Noisy Labels(LNL) by constructing a zero-shot classifier that can identify clean samples without relying on the in-training model itself. However, none of these methods are a technique for cleaning the labels of a dataset, and they are used just to select clean samples. In addition, none of the mentioned work addresses multi-label datasets and the methods can be used to address this issue. Hence, these methods are not suitable for presenting the quality of the dataset and capturing the problematic labels of a dataset.

In this work, we present a Vision-Language Sanitization and Refinement (VLSR) framework to sanitize the labels of large-scale datasets. Vision-language models are used because they excel in two key areas: (1) they are pretrained on large-scale, internet-sourced datasets which provides a vast and diverse understanding of the visual and textual world, and (2) they utilize a shared multi-modal embedding space which enables them to capture the relationship between textual and visual data through aligned extracted features. The CLIP model \cite{radford2021learning} is a state-of-the-art vision-language model that has a dual-encoder architecture with separate encoders for images and text based on transformers\cite{wolf2020transformers,ghiasvand2025few}. Transformers employ multi-head self-attention mechanisms to model complex relationships within sequences\cite{ahmadi2025unsupervised,khaniki2025class}. The text encoder processes tokenized input sequences and captures contextual nuances of words within a sentence. Similarly, the image encoder processes visual features extracted from input images\cite{adami2025gru,kermani2025systematic}. The resulting embeddings are mapped to a common latent space, allowing CLIP to measure the similarity between image-text pairs effectively. Below are the key contributions of the VLSR framework:

\begin{enumerate}
    \item \textbf{Dataset Sanitization}: The VLSR framework effectively sanitizes large-scale datasets by identifying the best and worst matches of image-label pairs. It detects misspellings, labels that do not correspond to the object in the image, and meaningless or irrelevant labels.
    \item \textbf{Handling Noisy Labels}: The VLSR framework is specifically designed to handle datasets with a large number of noisy labels, such as those extracted from web scraping or crowd-sourcing. It can group distinct but semantically similar labels, addressing the challenges posed by inconsistent or redundant label naming.
    \item \textbf{Multilabel Dataset Cleaning}: The VLSR framework extends its functionality to multilabel datasets by leveraging clustering and similarity scores. This approach enables the cleaning of labels even in cases where multiple labels are assigned to the same image, ensuring consistency and accuracy.
\end{enumerate}

\section{Methods}

The VLSR framework addresses the challenge of cleaning and organizing labels in large-scaled datasets. In this work, the aim is to clean the labels for a dataset containing $N$ images and $L$ distinct labels. The dataset is defined as $D = \{(I_i, L_i)\}_{i=1}^N$, where $I_i$ represents the $i$-th image, and $L_i = \{l_{i1}, l_{i2}, \dots, l_{ik}\}$ is the set of $k_i$ labels assigned to $I_i$. The total number of labels across all images is denoted by $L$, with $L = \bigcup_{i=1}^N L_i$. Given an image $I_i$ with multiple assigned labels $L_i$, the task is to produce a cleaned dataset $\hat{D} = \{(I_i, \hat{l}_i)\}_{i=1}^N$, where $\hat{l}_i$ is the single most appropriate label assigned to $I_i$. To achieve this goal, a label cleaning function $C$ was defined as:
\begin{equation}
    C(I_i, L_i) \rightarrow (I_i, \hat{l}_i),
\end{equation}
where $C$ takes an image $I_i$ and its associated labels $L_i$ as input and outputs the same image with a single, refined label $\hat{l}_i$. The refined label $\hat{l}_i$ is chosen from a subset of the original label space $\hat{L} \subseteq L$ such that $|\hat{L}| < |L|$. The CLIP model was utilized to generate embeddings for both images and their associated labels to effectively compute the $C$ and refine the noisy labels in our dataset.

\subsection{Generating Embeddings}

The CLIP model used in this work is pretrained on multiple multi-modal datasets and is effective in capturing semantic relationships between visual and textual data. It was employed to compute embeddings for each image $I_i$ and its associated set of labels $L_i = \{l_{i1}, l_{i2}, \dots, l_{ik}\}$. Let $\text{E}_{\text{img}}(I_i)$ denote the image encoding function and $\text{E}_{\text{text}}(l_{ij})$ denote the text encoding function of a label $l_{ij}$. These functions produce embeddings in a shared semantic space of dimension $d$, such that:
\begin{align}
    \mathbf{e}_i &= \text{E}_{\text{img}}(I_i) \in \mathbb{R}^d, \\
    \mathbf{e}_{ij} &= \text{E}_{\text{text}}(l_{ij}) \in \mathbb{R}^d \quad \forall l_{ij} \in L_i.
\end{align}

The embedding \( \mathbf{e}_i \) represents the semantic content of the image \( I_i \), while \( \mathbf{e}_{ij} \) represents the semantic meaning of the label \( l_{ij} \). These high-dimensional embeddings encode the relationships between visual and textual information and allow for a meaningful comparison between visual and textual information using geometric measures within the shared embedding space. In this work, cosine similarity was used as the geometric measure to quantify the alignment between images and labels.

\subsection{Dataset Sanitization with Cosine Similarity}

\begin{figure}[htbp!]
    \centering
    \includegraphics[width=0.95\linewidth]{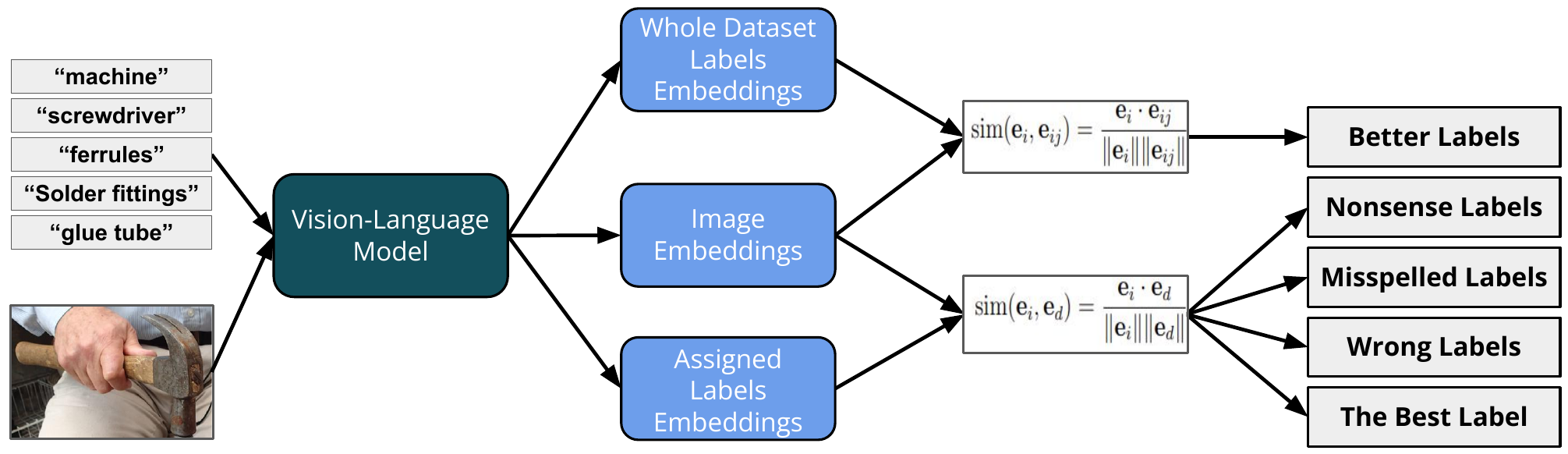}
    \caption{Overview of dataset sanitization method by the VLSR framework.} 

    \label{fig:TaskA}
\end{figure}

Figure \ref{fig:TaskA} illustrates the sanitization process. Image and label embeddings were generated using the CLIP model, followed by computing cosine similarity scores between image-label pairs.  Formally, the cosine similarity between two vectors \( \mathbf{a}, \mathbf{b} \in \mathbb{R}^d \) is defined as:
\begin{equation}
    \text{sim}(\mathbf{a}, \mathbf{b}) = \frac{\mathbf{a} \cdot \mathbf{b}}{\|\mathbf{a}\| \|\mathbf{b}\|},
\end{equation}
where \( \mathbf{a} \cdot \mathbf{b} \) is the dot product of the vectors, and \( \|\mathbf{a}\| \) and \( \|\mathbf{b}\| \) denote their Euclidean norms. A cosine similarity value of 1 indicates perfect alignment, while values closer to 0 suggest no meaningful similarity. Hence, instead of simply selecting the label with the highest similarity score for each image, these cosine similarity scores were used to conduct a deeper analysis into the semantic alignment between images and their assigned labels. This sorted information was used to verify the validity of existing labels, identify the strongest and weakest matches, and highlight potential errors in the labeling process. This process identifies issues such as labels that do not match the object, labels that are semantically nonsensical, and instances where entire scenes are mislabeled with overly specific terms. Furthermore, typographical errors, misspellings, or redundant variations in label text were detected. 

In this work, cosine similarity served as a diagnostic tool to improve dataset quality by comparing image-label pairs in two distinct approaches: image-to-assigned label and image-to-dataset comparisons. The image-to-assigned label comparison focuses on validating or rejecting the correctness of the labels originally associated with each image while the image-to-dataset comparison surfaces better matching labels that exist in the dataset but were not initially assigned.

\subsubsection{Image-to-Assigned Label Comparison}

In the first comparison, the embedding of each image \( \mathbf{e}_i \) was compared to each label embedding \( \mathbf{e}_{ij} \) in the set \( L_i \). This comparison evaluates how well each label corresponds to the content of the image:
\begin{equation}
    \text{sim}(\mathbf{e}_i, \mathbf{e}_{ij}) = \frac{\mathbf{e}_i \cdot \mathbf{e}_{ij}}{
    \|\mathbf{e}_i\| \|\mathbf{e}_{ij}\|}.
\end{equation}

A value of 1 indicates a perfect match, while values closer to 0 indicate a poor semantic alignment between the label and the image. Low cosine similarity values showcase the problematic labels, while high values indicate that the label perfectly matches the image.

\subsubsection{Image-to-Dataset Comparison}

In addition to comparing image embeddings with their associated labels, cosine similarity between the embedding \( \mathbf{e}_i \) of an image \( I_i \) and every label embedding in the entire dataset \( \mathcal{E}_L = \{\mathbf{e}_d \, | \, d = 1, 2, \dots, L \} \), where \( L \) is the total number of labels across the dataset were calculated. The similarity between \( \mathbf{e}_i \) and \( \mathbf{e}_d \) is computed as:
\begin{equation}
    \text{sim}(\mathbf{e}_i, \mathbf{e}_d) = \frac{\mathbf{e}_i \cdot \mathbf{e}_d}{\|\mathbf{e}_i\| \|\mathbf{e}_d\|}.
    \label{eq:6}
\end{equation}

A similarity value of 1 indicates a perfect match, while values closer to 0 indicate a poor semantic alignment between the label and the image. This approach provides a broader perspective by evaluating an image’s similarity to all labels in the dataset, rather than only its assigned ones. This score assists with identifying more semantically accurate labels that may already exist in the dataset but were not originally associated with the image.

\subsection{Clustering and Merging Labels}

In datasets collected from user inputs or web scraping, it is common to find many labels that look different but actually mean the same thing. These labels can carry minor differences in spelling, formatting, or phrasing, while they refer to the same concept. For instance, labels such as \textit{bicycle}, \textit{bike}, and \textit{bicycles}, or more specific variations like \textit{mountain bike}, \textit{road bicycle}, or \textit{kids' bike}, all describe the same general entity but are treated as distinct labels due to inconsistencies in representation. Labels can also vary in capitalization, the inclusion of special characters like underscores or dashes (e.g., \textit{mountain\_bike} or \textit{Mountain\_Bike} ), or minor misspellings (e.g., \textit{bicycl}) which do not alter the semantic meaning of a label but result in different representations within the dataset. Even more complex cases, such as \textit{electric bicycle} and \textit{e-bike}, highlight how different phrasing can convey the same meaning.

\begin{figure}[htbp!]
    \centering
    \includegraphics[width=0.95\linewidth]{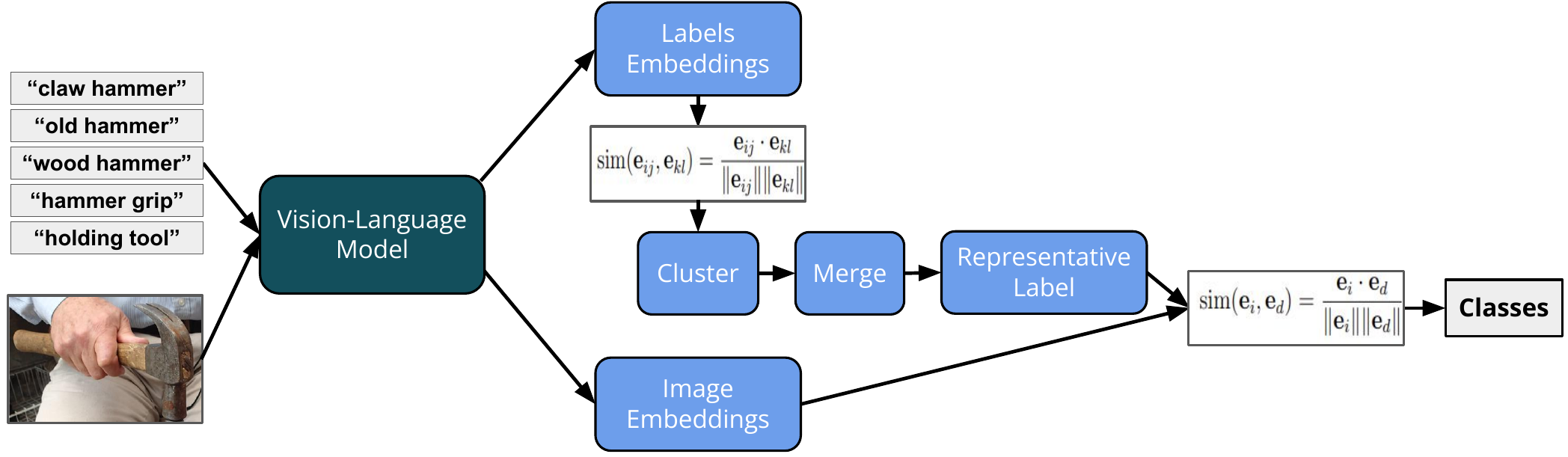}
    \caption{Overview of clustering and merging labels by the VLSR framework.}

    \label{fig:TaskB}
\end{figure}

To address these issues and by leveraging the fact that embeddings of similar words are positioned close to each other in the space, clustering methods were applied to group semantically or nearly equivalent labels together. The clustering of labels was performed using the DBSCAN (Density-Based Spatial Clustering of Applications with Noise) algorithm \cite{ester1996density}. DBSCAN is an unsupervised algorithm used for identifying clusters based on the density of data points in a region by grouping points that are closely packed together and marking points in low-density regions as noise. DBSCAN algorithm requires two key parameters: \( \varepsilon \) (eps), the maximum distance between two points to consider them neighbors, and the minimum number of samples, the minimum number of points required to form a dense region or cluster. Unlike traditional clustering techniques such as K-Means \cite{steinhaus1956division,lloyd1982least}, DBSCAN does not require prior specification of the number of clusters. This makes DBSCAN particularly suitable for this task since the exact number of semantically equivalent labels in the dataset is unknown. Figure \ref{fig:TaskB} represents the overview of the clustering and merging method.

Note that the DBSCAN algorithm typically calculates distances between points using Euclidean distance by default. However, in this task, cosine similarity was used as the measure to capture the semantic relationships in word embeddings, where the direction of the vectors holds more meaning than their magnitude. Hence, a cosine distance matrix was computed and provided as input to the DBSCAN algorithm. The cosine similarity between two label embeddings \( \mathbf{e}_{ij} \) and \( \mathbf{e}_{kl} \) was defined as:

\begin{equation}
    \text{sim}(\mathbf{e}_{ij}, \mathbf{e}_{kl}) = \frac{\mathbf{e}_{ij} \cdot \mathbf{e}_{kl}}{\|\mathbf{e}_{ij}\| \|\mathbf{e}_{kl}\|}.
\end{equation}
where \( \mathbf{e}_{ij}, \mathbf{e}_{kl} \in \mathbb{R}^d \) are the embeddings of labels \( l_{ij} \) and \( l_{kl} \) respectively. The cosine distance, based on this similarity, is defined as:

\begin{equation}
    \text{dist}(\mathbf{e}_{ij}, \mathbf{e}_{kl}) = 1 - \frac{\mathbf{e}_{ij} \cdot \mathbf{e}_{kl}}{\|\mathbf{e}_{ij}\| \|\mathbf{e}_{kl}\|}.
\end{equation}
A cosine distance of 0 indicates perfect similarity, while values closer to 1 suggest greater dissimilarity. Given the pairwise cosine distance matrix \( \mathbf{D} \), where each entry \( \mathbf{D}_{ij} \) represents the cosine distance between the label embeddings \( \mathbf{e}_{ij} \) and \( \mathbf{e}_{kl} \), the DBSCAN algorithm was applied using \( \mathbf{D} \). By setting appropriate values for \( \varepsilon \) and the minimum number of samples, DBSCAN identified clusters of labels that were semantically similar, while outliers or noise labels that did not fit into any dense cluster were left unclustered.

After clustering the labels, the frequency of each label's occurrence across the entire dataset was calculated. Within each cluster, the label with the highest frequency in the dataset was selected as the cluster's representative label. This approach ensures that the representative label is both semantically relevant and statistically dominant within the dataset. Once the representative labels were determined, all labels in the dataset that belonged to the same cluster were replaced with their respective cluster representative. 

Note that, among all the clusters, there were some with a low number of labels, which are considered noise or outliers. Clusters containing very few labels pose challenges for downstream tasks, as they do not provide sufficient examples for a classifier to learn meaningful patterns. To address this, a threshold-based merging strategy was implemented. This process involved identifying clusters with label counts below a predefined threshold and merging them into the closest neighboring cluster based on the cosine distance between their cluster representative label. The merging ensured that small clusters or outliers do not persist as isolated entities in the dataset. Merging small clusters reduces class imbalance by consolidating semantically similar labels, resulting in better representation across the dataset. 

The proposed VLSR method effectively addresses the multi-label nature of the dataset. As described earlier, each instance in the dataset is associated with multiple labels. After all the assigned labels were replaced with the cleaned representative labels from their clusters, another step was added to address the multi-label nature of the dataset. In this step, the cosine similarity between each representative label and its image was calculated using Eq.  (\ref{eq:6}). Among the possible labels for an image, the one with the highest similarity score was used as the final label. This ensures that each image is associated with the label that best matches its visual content.

\section{Results}

This section outlines the results of the experiments conducted to evaluate the VLSR framework. First, the dataset used for the experiments is introduced. Then, the process of generating embeddings is discussed. Then, the label sanitization using two different methods is outlined. Lastly, the process of clustering and merging the labels is discussed. 

\begin{figure}[htbp!]
    \centering
    \includegraphics[width=0.95\linewidth]{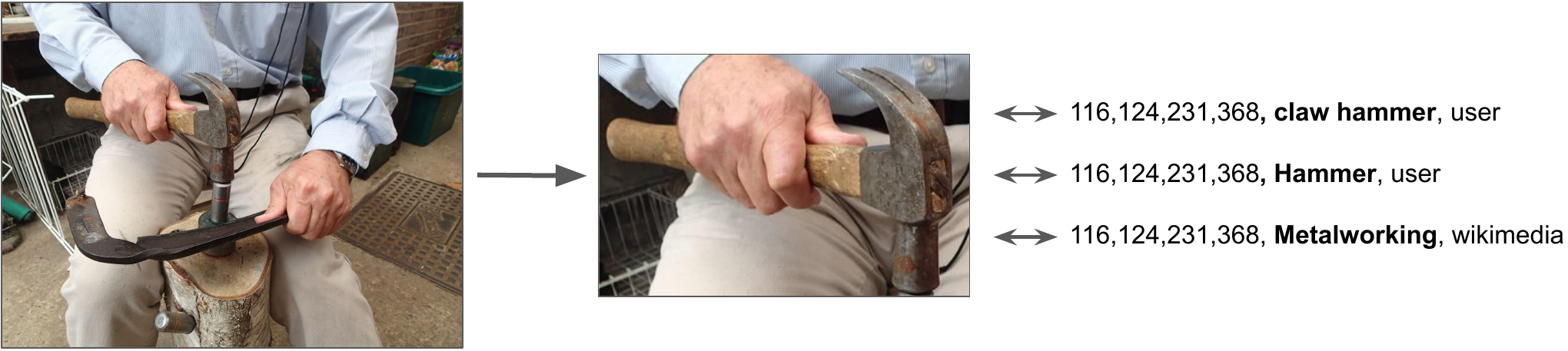}
    \caption{An example image from the Factorynet dataset. This example showcases bounding box annotations along with their associated labels and sources. Each entry includes the coordinates of the bounding box, the label (e.g., “claw hammer,” “Hammer,” “Metalworking”) for that bounding box, and the origin of the label.}

    \label{fig:Dataset}
\end{figure}

\subsection{Dataset}

The dataset used in the experiments is the Factorynet \cite{andrew_bowman_2024_13332887} dataset, which is a manufacturing industrial dataset containing a mix of human-generated and web-scraped labels. The dataset consist of 10,160 images and 6,426 distinct labels, while most of the labels exhibiting high error rates. Each image in the dataset is assigned multiple labels. This dataset is suitable for evaluating the effectiveness of the VLSR framework as it has a complete set of noisy labels. In this dataset, each image has a corresponding CSV file containing multiple labels and their labeling sources. All of the images and their assigned labels extracted form CSV files are the material used in the following experiments. Figure \ref{fig:Dataset} represents an example from the dataset.

\subsection{Generating Embeddings}

To analyze the dataset effectively, the first step involves generating embeddings. The embeddings generated in this process were 768-dimensional feature vectors produced by the OpenAI CLIP (vit-large-patch14) model from the Hugging Face Transformers library\cite{wolf2020transformers}. Two kinds of embeddings were generated: text embeddings and image embeddings. To generate the text embeddings, a file containing all unique labels was created as input for the CLIP model. Each label was tokenized using the CLIP tokenizer and passed through the model in text mode, and the resulting embeddings were saved in a binary format. The same procedure was applied to generate image embeddings, where each image was used as an input to the model in image mode. This process ensured consistency between the textual and visual embeddings. To preserve the meaning of labels such as "PPE" (personal protective equipment) or "2" wood screw" (containing special character) where altering the words could affect the context, no preprocessing method such as lowercasing or removing special characters was applied to the labels before applying the CLIP model.

\subsection{Dataset Sanitization}

\begin{figure}[htbp!]
    \centering
    \includegraphics[width=0.95\linewidth, height=6cm]{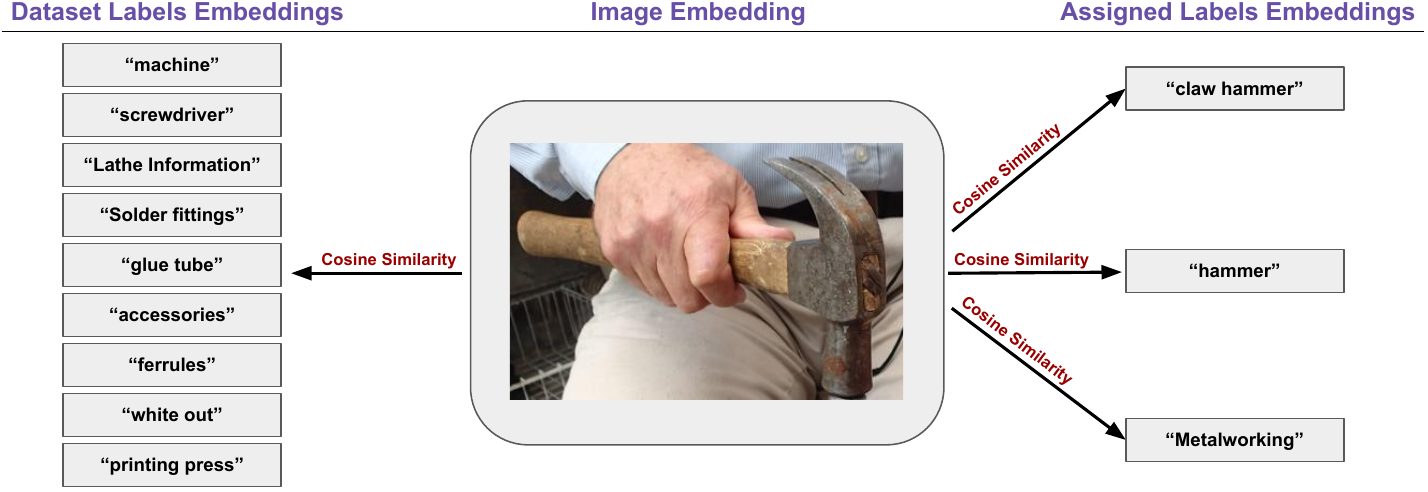}
    \caption{Example of cosine similarity comparison, where embedding of each image is compared with the embeddings of assigned labels and compared to all label embeddings in the dataset.}

    \label{fig:CosineSimilarity}
\end{figure}

After generating and saving the embeddings, the next step involved verifying the existing labels in the dataset. Figure \ref{fig:CosineSimilarity} illustrates an example of the data sanitization process. The image-to-assigned label and image-to-data analysis was conducted to result in the best sanitized label. The results of the dataset sanitization process showed significant improvements through both comparison strategies. 

\begin{figure}[htbp!]
    \centering
    \includegraphics[width=0.95\linewidth, height=20cm]{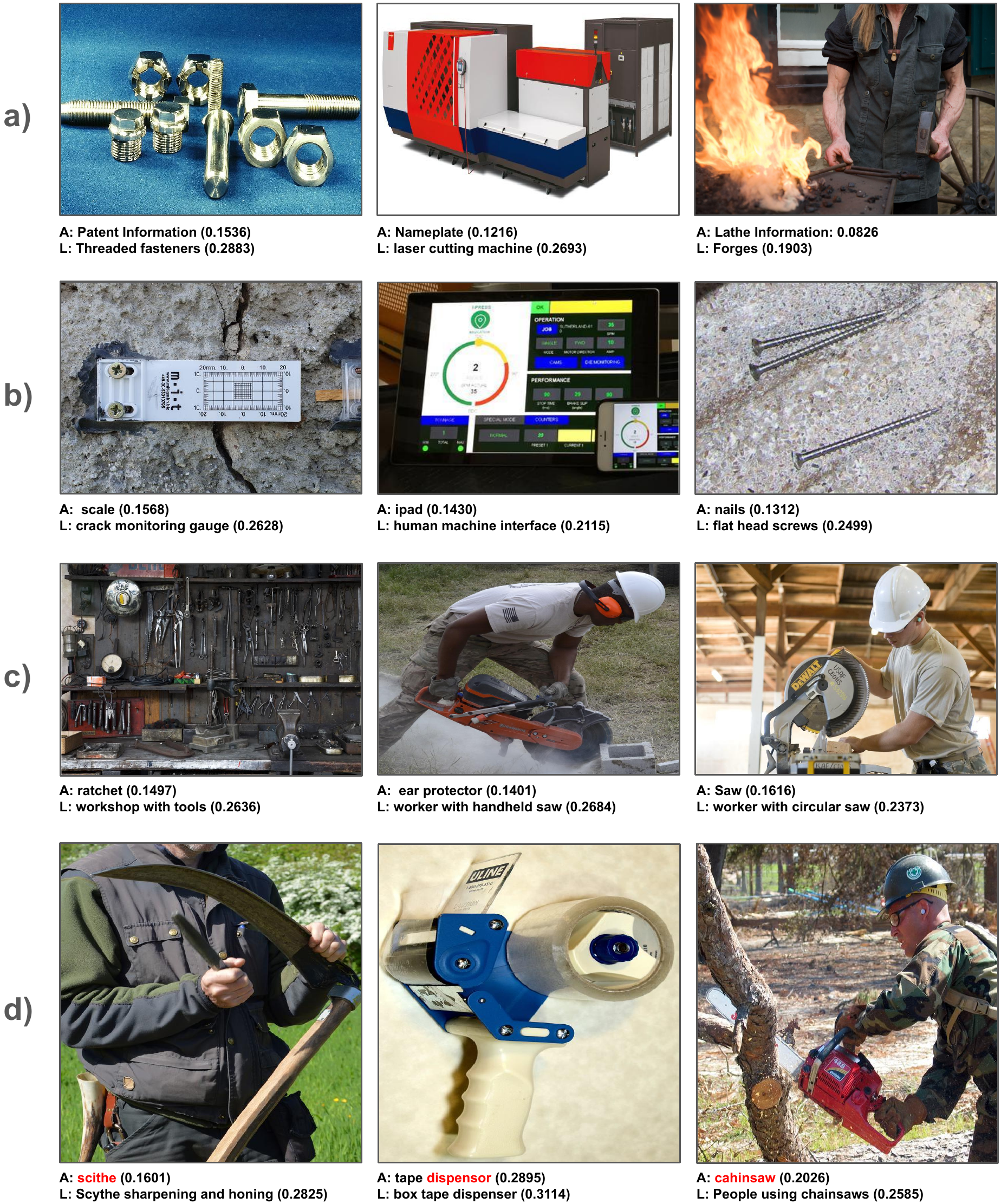}
    \caption{Examples of label sanitization results. Each example compares the originally assigned label (A) with the refined label (L) selected by the VLSR framework. The figure illustrates different error types, including: (a) \textit{Nonsense Labels}, (b)\textit{Incorrect Labels}, (c) \textit{Partial or Insufficient Labels}, and (d) \textit{Misspelled Labels} where (A) is wrong and (L) is correct. }

    \label{fig:Examples}
\end{figure}



The examples presented in Figure \ref{fig:Examples} illustrate several types of labeling errors that were identified through the dataset sanitization analysis. For each instance in Figure \ref{fig:Examples}, the originally assigned label (A) and its corresponding cosine similarity score with the image are shown alongside the refined label (L) selected by the VLSR framework and its similarity score. The first row highlights cases involving nonsensical or uninformative labels. The terms \textit{"Patent Information," "Nameplate,"} and \textit{"Lathe Information"} were repeatedly assigned to a huge number of images, while they do not represent the class of the objects. In contrast, the VLSR framework surfaced more semantically appropriate labels by retrieving the best-matching label from the entire dataset label set, and terms \textit{"Threaded fasteners," "laser cutting machine,"} and \textit{"Forges"} more accurately reflect the content of the corresponding images. 

The second row of Figure \ref{fig:Examples} showcases the incorrectly assigned labels. Terms such \textit{"scale," "Ipad,"} and \textit{"nails"} were assigned to the images that actually show a \textit{"crack monitoring gauge," "human machine interface,"} and \textit{"flat head screws"} as correctly identified by the VLSR framework. The third row of Figure \ref{fig:Examples} showcases where the assigned label (A) describes only a small part of the image rather than its main content. For instance, labels like \textit{"ratchet," "ear protector,"} and \textit{"saw"} refer to minor elements in the scenes, while the VLSR framework correctly identified broader and more accurate labels such as \textit{"workshop with tools," "worker with handheld saw,"} and \textit{"worker with circular saw,"} which better reflect the full context of each image. 

The last row of Figure \ref{fig:Examples} illustrates examples where the framework successfully detected misspelled labels. The assigned labels like \textit{"scithe," "tape dispensor,"} and \textit{"cahinsaw"} carry spelling errors, and the VLSR framework correctly assigned them to \textit{"Scythe sharpening and honing," "box tape dispenser,"} and \textit{"People using chainsaws,"} respectively.

\begin{figure}[htbp!]
    \centering
    \includegraphics[width=0.95\linewidth, height=9cm]{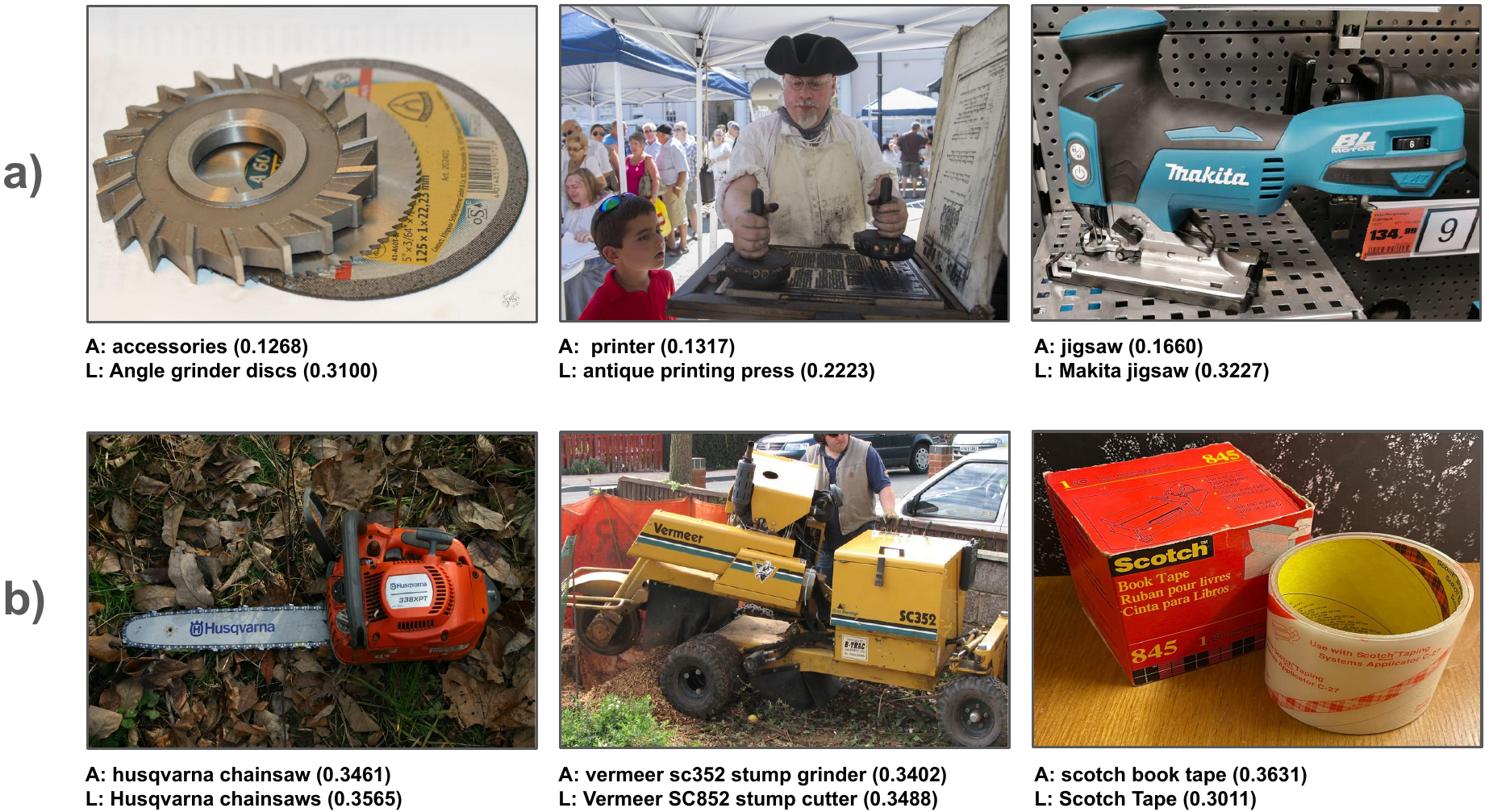}
    \caption{Examples of label sanitization results. Each example compares the originally assigned label (A) with the refined label (L) selected by the VLSR framework. The figure showcases the ability of the framework to reveal correct assigned labels and suggest another label with minimal difference such as (a) \textit{Improved Matches} where A and (b) \textit{Best Matches} where both (A) and (L) are valid but (L) offers better specificity.}

    \label{fig:BestExamples}
\end{figure}

The VLSR framework can also identify more precise or descriptive labels even when the original labels are technically correct. This is demonstrated in the first row of Figure \ref{fig:BestExamples}, where the assigned labels \textit{"accessories," "printer,"} and \textit{"jigsaw"} are valid, but the framework selected more specific alternatives like \textit{"Angle grinder discs," "antique printing press,"} and \textit{"Makita jigsaw"} based on higher similarity scores, which describe the image content with a more detailed representation. The second row of Figure \ref{fig:BestExamples} demonstrates how the VLSR framework can confirm the correctness of assigned labels when they closely match the image content. In these examples, labels such as \textit{"husqvarna chainsaw," "vermeer sc352 stump grinder,"} and \textit{"scotch book tape"} are validated by the framework, which selects nearly identical alternatives as \textit{"Husqvarna chainsaws," "Vermeer SC852 stump cutter,"} and \textit{"Scotch Tape"} with only minor differences in casing or phrasing.

The outputs from these two processes provide actionable approaches into how to address errors in the dataset. Depending on the dataset's characteristics and features, different strategies can be employed to resolve these issues. Labels with consistently low similarity scores that do not meaningfully describe any image can be removed entirely, such as "Lathe Information". For datasets where human intervention is feasible, problematic labels can be manually reviewed and corrected. A threshold-based approach can be applied to automate label adjustments. For example, if the similarity score between an image and its assigned label is lower than its similarity score to another label in the dataset, the assigned label can be replaced with the better match.
This framework significantly reduces the effort required for manual labeling and verification, which are traditionally time-consuming and costly processes. By identifying and addressing errors effectively, it ensures higher-quality datasets, enabling more accurate and reliable downstream applications.

\subsection{Clustering and Merging Labels}

To perform clustering, text embeddings generated by the CLIP model for all labels were used as input for the DBSCAN algorithm. Multiple configurations of the clustering algorithm were tested by varying the values of the \( \varepsilon \) and the number of samples parameters to evaluate the impact of different clustering behaviors on the data. After clustering the labels, some clusters contained very few samples and were considered likely to be noise or overly specific label variants. In order to further reduce the noise, these small clusters were iteratively merged into the closest larger clusters based on cosine distance between their centroid embeddings. The most frequent label in each cluster was then selected as the representative label for all labels in that group.


The Factorynet dataset with its multi-label assignments for each image and numerous noisy labels exhibits a low image-to-label ratio. A large number of distinct labels were assigned to different images, and many of these labels essentially carry the same meaning, referring to the same concept. While humans can easily recognize the semantic similarity between these labels, a machine treats each label as a distinct class. This actually increases the complexity of the learning process and needs to be fixed. For instance, the words \textit{"box tape dispenser," "tape dispenser," "Handheld tape dispensers,"} and \textit{"tape dispensor"} all refer to the same object class but vary due to misspellings and redundant detail. As another example, labels such as \textit{"PPE"} and \textit{"Personal Protective Equipment"} convey the exact same meaning but are treated as distinct classes from the model’s perspective. 

In the experiments with the Factorynet dataset, higher values of \( \varepsilon \) (e.g., 0.1) and the minimum number of samples (e.g., 2) resulted in the formation of large clusters that grouped together labels with significantly different meanings. For instance, labels such as \textit{"Electronic locks"}, \textit{"automated dispenser"}, \textit{ball lock}, \textit{"triangle files"}, \textit{"Minting punches"}, \textit{"glass jars"}, and \textit{"bell"} were incorrectly clustered into the same group, despite referring to conceptually unrelated objects. This result highlights the risk of semantic drift when clustering parameters are not appropriately constrained. After evaluating multiple configurations, the most coherent groupings of labels that shared a clear semantic relationship were achieved with the \( \varepsilon \) = 0.07 and minimum number of samples set to 1. Table \ref{tab:clusters} presents examples of representative clusters formed from this process, demonstrating successful grouping of labels that differed due to formatting, misspellings, or overly specific phrasing but shared the same semantic meaning. Each cluster groups together labels that, despite differences in casing, formatting, spelling, or level of specificity, refer to the same underlying concept. For example, in Cluster 8032, various helmet-related labels such as \textit{"helmet with face shield and earmuffs," "safety helmet with earmuffs,"} and \textit{"ABUS bicycle helmets"}are grouped under the representative label \textit{"helmet."} This demonstrates the effectiveness of the method in identifying and consolidating semantically similar labels. These results illustrate how the VLSR framework successfully reduces label noise and redundancy by leveraging semantic similarity, contributing to improved dataset quality and downstream model training. The method successfully reduced the total number of labels in the whole dataset from 6,426 to 408 distinct labels. The selection of \( \varepsilon \) and the minimum number of samples remains a sensitive decision that directly affects clustering quality. At present, tuning these hyperparameters requires human inspection of cluster outputs to ensure semantic integrity. While the process is semi-automated, full automation—potentially involving a secondary model to validate cluster cohesion is a direction for future work.

\begin{table}[H]
\centering
\scriptsize  
\caption{Sample Clusters Generated Using DBSCAN with \( \varepsilon \) = 0.07 and the minimum number of samples set to 1. Each row with a \textbf{Cluster ID} shows a cluster identified from the CLIP-generated text embeddings. The \textbf{Rep. Label} is the cluster's representative label and is the most frequent label in the dataset. \textbf{Rep. Freq.} indicates its occurrence in the dataset, while \textbf{Total Freq.} denotes the combined frequency of all labels in the cluster. \textbf{Example Labels} demonstrates how labels with different casing, misspellings, or redundant detail have been successfully grouped together under a unified concept.}
\label{tab:clusters}
\begin{adjustbox}{width=\textwidth}
\begin{tabularx}{\textwidth}{|c|l|c|c|X|}
\hline
\textbf{Cluster ID} & \textbf{Rep. Label} & \textbf{Rep. Freq.} & \textbf{Total Freq.} & \textbf{Example Labels} \\
\hline
8032 & helmet & 34 & 56 & helmet with face shield and earmuffs, helmets, safety helmet with ear muffs, Helmet, Abus helmet, helmet, safety helmet with earmuffs, helmet with facemask, ABUS bicycle helmets, helmet with face shield, cap, safety helmet with earmuffs and face shield, safety helmet with face shield and earmuff \\
\hline
8330 & Jigsaws & 16 & 35 & Bosch jigsaws, makita jigsaw, Jigsaws, Disassembled jigsaws, jigsaw, Makita jigsaws, Jigsaw, Ryobi jigsaws, Makita jigsaw, mounted jigsaw, skill jigsaw \\
\hline
8456 & Saws & 32 & 100 & back saw, Scroll saws, Gang saws, Saws, Pad saws, Back saws, Sabre saws, Saws in the Veenpark, Saws in China, Cold saws, Ice saws, Pit saws, saws, cold saw, Sawyers, Milwaukee sabre saws, Saws in the United States, Coping saws \\
\hline
8468 & corkscrew & 8 & 47 & Corkscrew (tool) - wing type, cork screws, Pocket corkscrews, Figurative corkscrews, antique corkscrew, vintage corkscrew, Corkscrew, cork screw, corkscrew, Simple full metallic corkscrews, Corkscrew (tool) - lemonade type, Corkscrew (tool) - spiral wine opener, Corkscrew (tool) - basic type, Rack-assisted corkscrews, Corkscrews by museum \\
\hline
8550 & windmill & 28 & 55 & wind mill, Windmills, windmill in field, windmills, windmill, Windmill \\
\hline
8486 & Screwdrivers & 79 & 111 & screwdrivers, Ratchet screwdrivers, Yankee screwdrivers, Screwdrivers, Phillips screwdrivers, Electric screwdrivers, Jeweler's screwdrivers, Electrician screwdrivers, flat head screwdrivers, precision screwdrivers, Stanley electric insulated screwdrivers, Flat head screwdrivers, Stanley screwdrivers \\
\hline
8573 & wrench & 10 & 21 & wrenches, metal wrenches, allen wrenches, Wrenches, wrench set, wrench, metal wrench set \\
\hline
8669 & belt sander & 13 & 20 & combination belt and disc sander, handheld belt sander, belt sander, horizontal belt sander, vertical belt sander, flott ksm 150 cross belt sander \\
\hline
8707 & Hinges & 12 & 44 & Hinges in the United States, Door hinges, Concrete hinges, Overhead cabinet hinges, Hinge, Laptop hinges, Hinges, Window hinges, door hinge, Game \& Watch (hinges), hinge, Cup hinges, Door hinges in the United States, decorative hinge \\
\hline
8770 & welding power generator & 7 & 26 & arc machines inc welding generator, miller syncrowave 350 lx welding power generator, welder generator, welding power generator, welding generator, Welding power supplies, fronius wtu 657 welding generator, welding power supply, jackle plasma 110i plasma welder generator \\
\hline
8806 & Laser cutting & 11 & 45 & laser cutting head, laser engraver, laser cutter, hankwang laser cutting machine, cnc laser cutter, Laser engraving, CNC laser cutter, laser cut wood router, Laser-cut products, laser, Laser cutting, laser cutting machine, hankwang ps6020 laser cutter, Laser cutters, gravograph ls800 laser engraver, enclosed laser cutter, epilog laser cutter \\
\hline
8849 & Padlocks & 12 & 56 & Padlock textures, Bicycle O-locks, abloy 330 padlock, Bicycle chain locks, Square photos of padlocks, padlock, Bicycle locks, Bicycle rear derailleur hangers, abloy padlock, Padlocks with chains, bike lock, Quality images of padlocks, padlock with cover, Padlocks, Love padlocks, Bicycle cable locks, Bicycle U-locks, Rusty padlocks in Canada, Padlock, Combination padlocks, krypto 10k bike lock, smart padlock\\
\hline
8923 & metal sheets & 1 & 24 & sheet metal, thick metal sheets, sheet metal brake, zeziola dtz-130 sheet metal former, rolled metal sheet, metal shims, Steel slabs, press brake machine, stack of thick metal sheets, metal slabs, metal sheets, Guillotines (sheet metal), stack of metal sheets, metal machine wall, sheet metal former, metal blades, metal slab\\
\hline
\end{tabularx}
\end{adjustbox}
\end{table}



\section{Conclusion}

This paper presents a VLSR framework for refining and sanitizing labels in large-scale, multi-label image classification datasets. Both the images and their associated labels are embedded into a shared semantic space using CLIP vision-language model to evaluate the label quality and correctness. Two complementary tasks are addressed by calculating cosine similarity between the embeddings generated by CLIP model. First, label sanitization was performed to identify problematic labels such as those that are misspelled, irrelevant to the image, or semantically weak in describing the object class. This was achieved by comparing the cosine similarity score between each image and all of its assigned labels. Additionally, the framework surfaces more semantically appropriate label alternatives by computing similarity between each image and all labels in the dataset. Second, a density-based clustering method was applied to the label embeddings to group semantically similar labels. This addresses issues such as inconsistent casing, formatting, synonymy, misspellings, and overly specific or ambiguous descriptors. To further improve the label coherence, an iterative merging step was then applied to consolidate small or redundant clusters. Experiments conducted on the Factorynet dataset, which contains over 10,000 images and 6,000 labels sourced from crowdsourcing and web scraping, demonstrate the effectiveness of the VLSR framework. Qualitative examples show the framework’s ability to identify and correct labeling issues with minimal human intervention and significantly lower labeling costs. The resulting clusters showcase the method’s strength in unifying label semantics and improving dataset quality.

The proposed approach is generalizable and scalable, making it applicable to a wide range of domains where label noise is prevalent. Future work will explore automating the sanitization process and enabling fully unsupervised identification and correction of noisy labels through adaptive thresholding of cosine similarity scores. Furthermore, foundation models may be employed to validate the coherence of each cluster by verifying the semantic relatedness of the grouped labels with a dynamic adjustment of clustering hyperparameters, such as epsilon and the minimum number of samples, to reduce clustering errors and improve label consistency. Lastly, future work could explore extending the VLSR framework to other modalities, such as video or 3D data, and incorporating external knowledge bases or weak supervision signals to enhance semantic understanding and label accuracy.

\section*{Authorship Credits}
\textbf{Nazanin Mahjourian}: Methodology, Algorithm Development, Execution, Visualization, Analysis, Original Drafting.\textbf{Vinh Nguyen}: Methodology, Analysis, Review and Editing

\section*{Acknowledgment}
The authors declare they have no conflict of interests that could have appeared to influence the work reported in this paper.

\section*{Funding Resources}
This research was supported by Michigan Technological University Department of Mechanical Engineering - Engineering Mechanics.

\appendix


\bibliographystyle{elsarticle-num} 
\bibliography{LabelCleaning}

\end{document}